\documentclass[letterpaper, 10 pt, conference]{ieeeconf}  %

\IEEEoverridecommandlockouts                              %

\usepackage{graphics} %
\usepackage{epsfig} %
\usepackage{mathptmx} %
\usepackage{times} %
\usepackage{amsmath} %
\usepackage{amssymb}  %

\usepackage{bm}
\usepackage[dvipsnames]{xcolor}
\usepackage[breaklinks=true,colorlinks=true,linkcolor=red,citecolor=red]{hyperref}

\usepackage{booktabs}
\usepackage{multirow}
\usepackage{soul}
\usepackage{bm}

\title{\LARGE \bf
Multi-Finger Grasping Like Humans
}

\author{Yuming~Du$^1$, Philippe~Weinzaepfel$^2$, Vincent~Lepetit$^1$ and Romain~Br\'egier$^2$%
\thanks{$^1$ Y.~Du (\texttt{yuming.du@enpc.fr}) and V.~Lepetit are with LIGM, Ecole des Ponts, Univ Gustave Eiffel, CNRS, Marne-la-Vall\'ee, France.}%
\thanks{$^2$ P.~Weinzaepfel and R.~Br\'egier are researchers at NAVER LABS Europe, Meylan, France. \url{https://www.europe.naverlabs.com}.}
}

\begin{document}

\maketitle
\thispagestyle{empty}
\pagestyle{empty}

\begin{abstract}
Robots with multi-fingered grippers could perform advanced manipulation tasks for us if we were able to properly specify to them what to do.
In this study, we take a step in that direction by making a robot grasp an object \emph{like} a grasping demonstration performed by a human.
We propose a novel optimization-based approach for transferring human grasp demonstrations to any multi-fingered grippers, which produces robotic grasps that mimic the human hand orientation and the contact area with the object, while alleviating interpenetration.
Extensive experiments with the Allegro and BarrettHand grippers show that our method leads to grasps more similar to the human demonstration than existing approaches,
without requiring any gripper-specific tuning.
We confirm these findings through a user study and validate the applicability of our approach on a real robot.

\end{abstract}

\section{INTRODUCTION}

\begin{figure}
\centering
\includegraphics[width=1.0\linewidth]{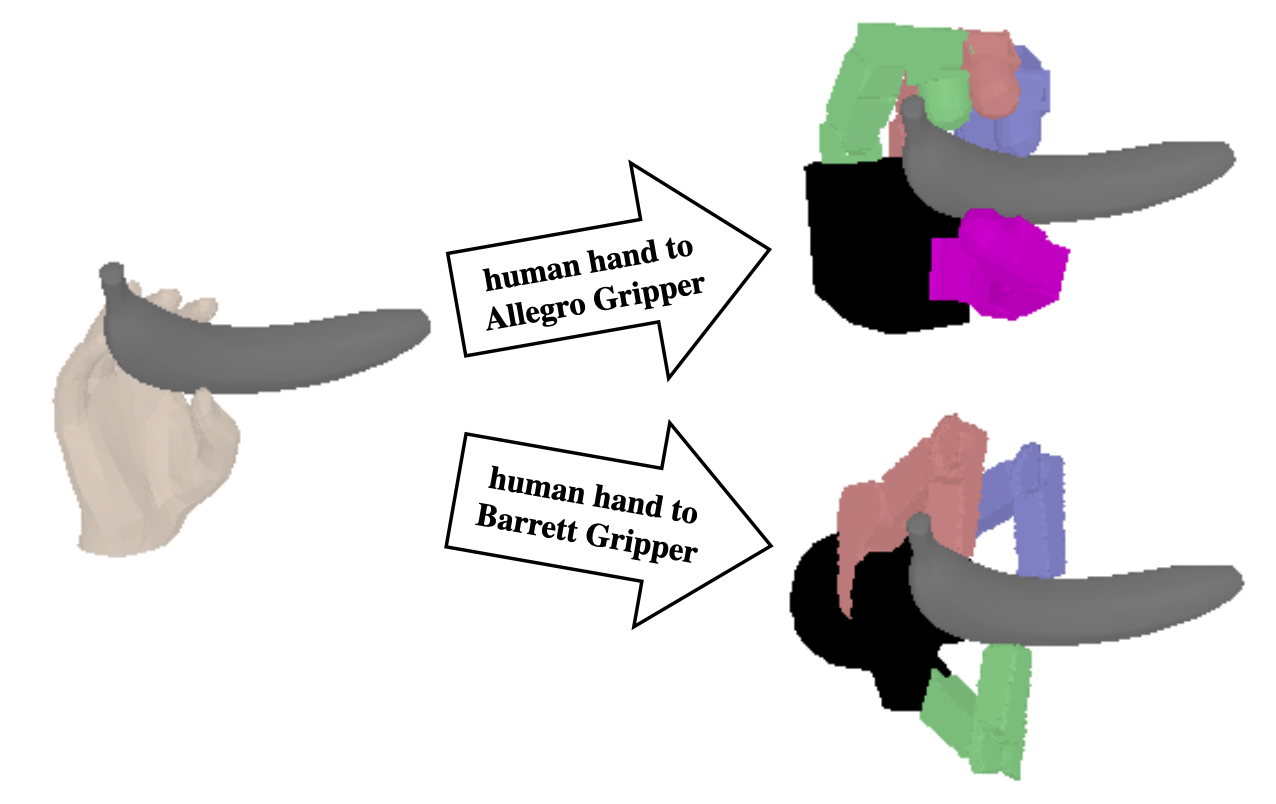}
\vspace{-0.7cm}
\caption{\textbf{Grasping like humans.} Given an input human grasp (left), our method outputs a configuration of a multi-fingered gripper grasping the same object %
\emph{like the human} demonstration. We experiment with the Allegro (top) and BarrettHand (bottom) grippers.
}
\label{fig:introduction} 
\end{figure}

\begin{figure}
\centering
\includegraphics[width=0.8\linewidth]{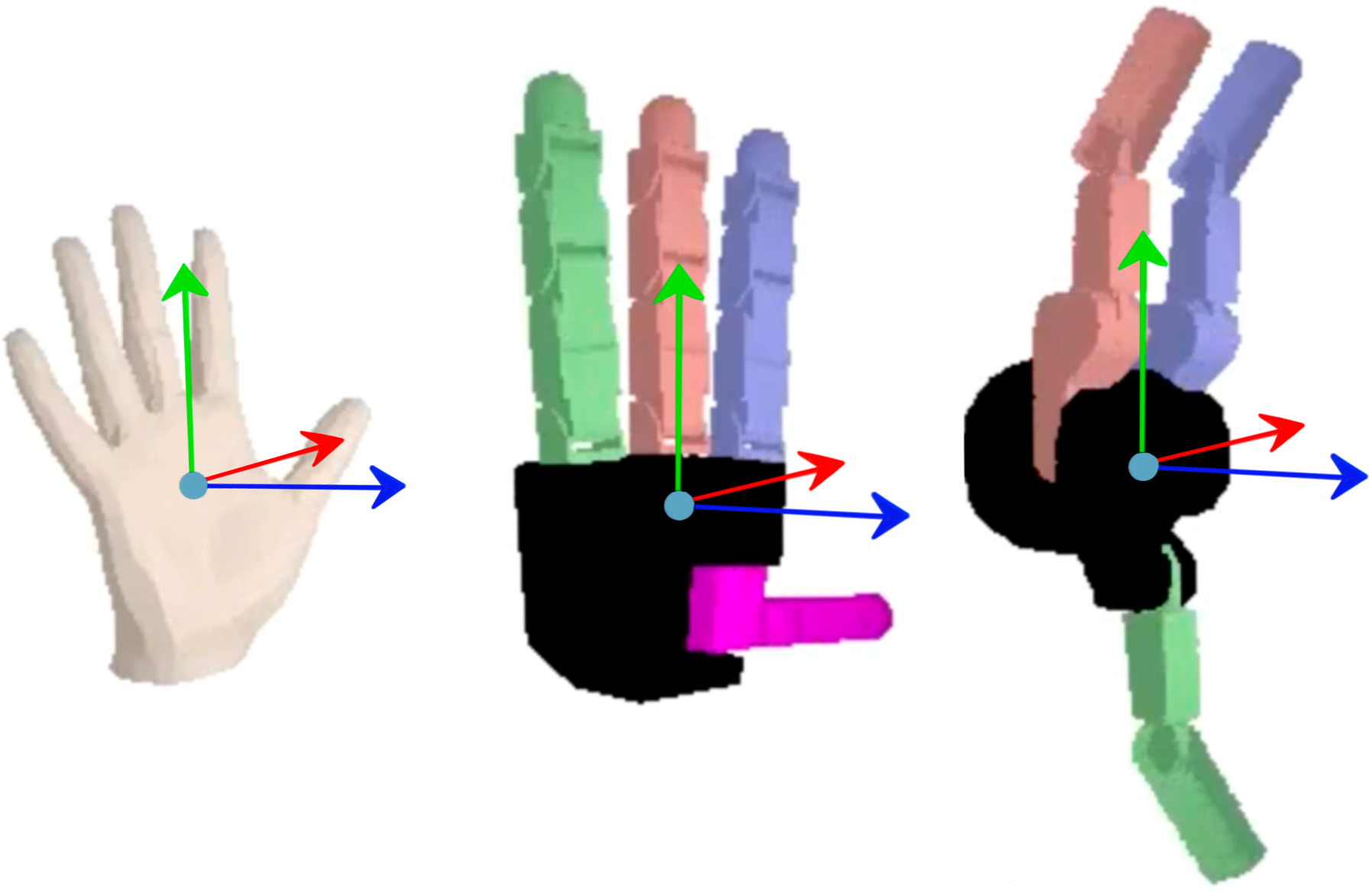}
\vspace{-0.5cm}
\caption{\textbf{Comparison between different grippers} at the same scale with a human hand (left), Allegro (middle) and BarrettHand (right). Note that the size of the gripper and in particular the fingers are significantly different. The \textcolor{blue}{blue} vector represents the normal vectors of the human hand and robot hands, the \textcolor{green}{green} vector represents the forward vector (best seen in color).}
\label{fig:comparison}
\end{figure}

To assist people in their daily activities, a robot would need to manipulate objects in specific ways, dependent on its current task, \emph{e.g.} it should not handle a knife in the same way for cutting vegetables as for handing it to a person with reduced mobility.
A human could teach such specificities to a robot by demonstrating the task to reproduce.
In this work, we focus on the task of grasping rigid objects, as illustrated in Figure~\ref{fig:introduction}. 
Reproducing exactly a human grasp is impossible for a robot, because a robotic gripper is usually quite different from a human hand: different size, number of fingers, actuation, \emph{etc.} (see Figure~\ref{fig:comparison} for a comparison).
Instead, the robot should grasp objects \emph{like} the human did.

Most existing grasp retargeting approaches~\cite{dexpilot,dexmv,contacttransfer} rely on handcrafted correspondences between the human hand and the robotic gripper, either in terms of joints, surfaces, or key vectors, and they do not consider the object. 
Some other methods such as ContactGrasp~\cite{contactgrasp} refine the grasps produced by GraspIt!~\cite{graspit} by optimizing the contact surface and reranking them. However, such approach is slow as it takes tremendous time and effort to generate and refine grasp candidates. It can moreover lead to significant differences with the human grasp.

The fundamental problem is that the exact meaning of ``grasping like a human'' is not well defined.
In this study we nonetheless introduce some generic proxies for grasp similarity, namely the \emph{contact} surface and the grasp \emph{orientation}.
Indeed, the affordance~\cite{gibson1979ecological} of a grasped object is typically dependent on the open space surrounding this object. Grasping an object from a similar orientation, with a similar contact surface on the object as in the human demonstration should therefore in general enable to perform with this object similar actions as the human.
In this paper, we investigate how well the use of these proxies enable to produce robotic grasps similar to human demonstrations.

To do so, we propose a multi-step optimization-based method that takes a human grasp demonstration as input -- represented by a 3D mesh of the object and a parametric MANO model of the hand pose~\cite{mano} -- and returns the configuration of the corresponding robotic grasp.
We define an objective function that encourages similar contact surfaces and global orientation for the human and the robotic grasps, while penalizing interpenetrations of the gripper and the object.
To avoid local minima, we perform a multi-stage optimization where the gripper global position and orientation are initialized similarly to the human demonstration.
Fingers are then closed by minimizing the distance between the fingertips and contact areas on the object, before optimizing for our full objective function in a last step.
To validate the genericity of our approach, we experiment with two off-the-shelf robotic hands: the \emph{Allegro}~\cite{allegrohand} and the \emph{BarrettHand}~\cite{barretthand} grippers (see Figure~\ref{fig:comparison}).

We evaluate our method using human grasps from the YCB-affordance dataset~\cite{ganhand} with various quality metrics, and we also perform a user study to compare qualitatively our approach with related methods. Both evaluations show that our approach allows to predict reasonable grasps that are better -- and more similar to the human demonstration -- than existing state-of-the-art grasp retargeting methods.
In the end, we validate the applicability of the approach in the real world on a Panda robotic arm.

In summary, the main contributions of this work are: (1) A novel objective function consisting of four losses which encourages a valid grasp while capturing the similarity between the human hand grasp and robot gripper grasp. (2) A novel multi-step optimization-based pipeline to transfer a human grasp demonstration to any multi-fingered gripper. (3) An extensive quantitative and qualitative evaluation and comparison between our approach and other related methods.

\section{Related work}

In this section, we review various works related to grasping like humans with multi-fingered grippers.

\paragraph*{Grasp prediction}
Predicting potential grasps for a given object is a classical research topic, illustrated by the seminal \emph{GraspIt!} simulator~\cite{graspit}.
Most recent approaches~\cite{multifingan,wu2020generative,varley2015generating,lu2020multi,liu2019generating} focus on learning-based techniques,
with some approaches~\cite{contactgraspnet,wu2020generative} modeling reachability constraints in the scene.

\paragraph*{Grasping from demonstration}
Learning from demonstration is also an important paradigm in robotics~\cite{pomerleau1991efficient,billard2008survey,argall2009survey,rajeswaran2017learning}.
It aims at teaching a particular task to a robot from a few examples of a human performing a similar task.
Most current approaches for learning from demonstration in the context of object manipulation focus on complex manipulation tasks with simple parallel-jaw grippers~\cite{zhang2018deep,schmeckpeper2020reinforcement}. 
On the contrary, we focus in this study on simpler manipulation tasks %
(static grasping)
but with more complex multi-fingered grippers -- that could allow more advanced grasps and manipulations.

\paragraph*{Pose retargeting}
A solution to transfer a human grasp demonstration to a robotic gripper it to define some fixed correspondences between the human hand and the robotic gripper. DexPilot~\cite{dexpilot} and DexMV~\cite{dexmv} use some handcrafted motion retargeting techniques to do so.
Similarly, ContactTransfer~\cite{contacttransfer} relies on fixed correspondences between the surface of the human hand and the robotic gripper.
The applicability of such approaches is arguably limited however because the human hand and the gripper may have significantly different characteristics in practice. 
For instance, the Allegro gripper has only 4 fingers and is about 1.6 times larger than a typical human hand (see Figure~\ref{fig:comparison}).
Moreover, these approaches do not consider the object being grasped.

\paragraph*{Contact-based retargeting}
More related to our work are approaches trying to predict robotic grasps sharing similar contact areas with the object as in the human demonstration, without requiring explicit correspondences between fingers of the human and the robot.
In~\cite{zhu2021toward}, Zhu~\emph{et al.} propose to annotate functional parts of the objects -- \emph{i.e.}, where humans would grasp the object or not -- to generate potential grasps for these objects.
Recently, ContactGrasp~\cite{contactgrasp} was proposed and uses GraspIt!~\cite{graspit} to generate a set of grasps that are iteratively refined and reranked such that the contact areas of the gripper on the object become closer to the ones of the human grasp. This approach has several drawbacks however.
First, it is about 40 times slower than our method as it has to generate and refine hundreds of grasp candidates each time.
Second, \emph{GraspIt!} mainly generates power grasps, and thus the refined grasps have similar properties.
Third, by focusing only on contact areas, ContactGrasp can produce grasps in which the gripper is occluding some important parts for the affordance of the grasped object. 
In comparison, our proposed optimization approach is faster 
and leads to grasps more similar to human ones, thanks to a simple yet effective initialization and thanks to additionally taking into account the grasp orientation. %

\section{GRASPING LIKE HUMANS}
\label{sec:optim}

In this section, we describe our optimization-based approach to generate a robot grasp `similar' to a given human grasp. %
After formalizing the problem and notations (Section~\ref{sub:problem}), we introduce the optimized objective function in Section~\ref{sub:loss} and detail all the steps of our approach in Section~\ref{sub:pipeline}.

\subsection{Problem and Notations}

\label{sub:problem}

We consider as input a rigid object grasped by a human hand. We represent the object by a 3D mesh $\mathcal{M}_{object}$, and we adopt the MANO~\cite{mano} model to represent the pose of the hand by a global rigid transformation $(R_{hand}, t_{hand}) \in SO(3) \times \mathbb{R}^3$ relative to the object and by its local joints configuration $\theta_{hand} \in SO(3)^{20}$.
Similarly, we assume that a kinematic model of the robotic gripper is available.
We aim to predict a global pose $(R_{robot}, t_{robot}) \in SO(3) \times \mathbb{R}^3$ relative to the object and some joints configuration $\theta_{robot} \in \mathbb{R}^n$ describing a static grasp with this gripper similar to the human demonstration ($n=16$ for Allegro, $n=7$ for Barrett).
We formulate this as an optimization problem and  minimize an objective function $\mathcal{L}(R_{robot}, t_{robot}, \theta_{robot})$ representing the \emph{dissimilarity} of the robotic grasp with the human demonstration.

\begin{figure}
\center
\includegraphics[width=0.9\linewidth]{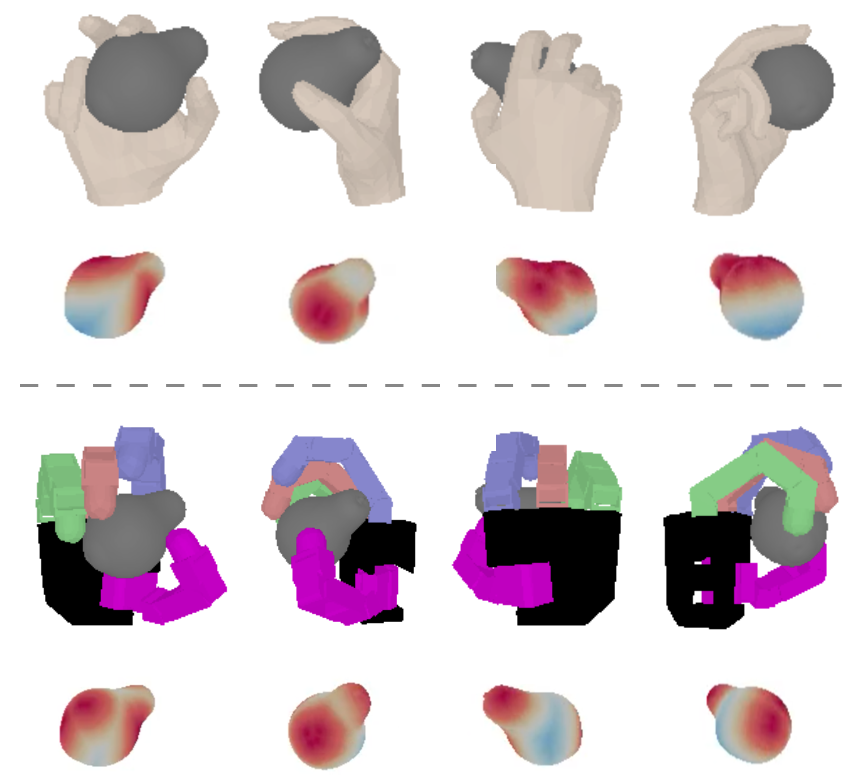}
\vspace{-0.3cm}
\caption{\textbf{Contact heatmaps} on the object mesh corresponding to a human (top) and robotic (bottom) grasp. Our optimization-based approach tries to minimize the discrepancy between these heatmaps. 
Red color denotes regions close to the hand/gripper while blue color denotes regions far from the hand/gripper.
}
\label{fig:contact-region-heatmap}
\end{figure}

\subsection{Objective Function}
\label{sub:loss}

Our objective function $\mathcal{L}$ is composed of a contact-heatmap loss $\mathcal{L}_{C}$ that incites contacts on the object to be similar, a hand orientation loss $\mathcal{L}_{O}$, as well as losses that penalize  interpenetration with the object $\mathcal{L}_{I}$ and self-penetration of the gripper $\mathcal{L}_{S}$, \emph{i.e.}:
\begin{equation}
\label{eq:objective_function}
      \mathcal{L} = \lambda_{C} \mathcal{L}_{C} +  \lambda_{O} \mathcal{L}_{O} + \lambda_{I} \mathcal{L}_{I} + \lambda_{S} \mathcal{L}_{S}
\end{equation}
with weights experimentally set to $\lambda_{C}=10, \lambda_{O}=10, \lambda_{I}=0.5$ and $\lambda_{S}=1$. We detail these terms in the following paragraphs.

\paragraph{Contact Heatmap Loss $\mathcal{L}_C$}

Intuitively, grasps are similar if their contact regions on the target object are similar. Based on this observation, we propose an object-centric contact heatmap loss term, which encourages the contact regions of the input human hand and the robotic gripper on the object to be similar.
Specifically, we represent the contact regions of the human hand and the robot gripper by scalar contact heatmaps $H$ on the object.
At each vertex $o_i \in \mathcal{M}_{object}$ of the object mesh, we define the values of these heatmaps as
\begin{equation}
\left\lbrace
\begin{aligned}
     H^{hand}(o_i) &= \exp(-d(o_i, \mathcal{M}_{hand})/\tau) \\
     H^{robot}(o_i) &= \exp(-d(o_i, \mathcal{M}_{robot})/\tau) \> 
 \end{aligned}
 \right.
\end{equation}
where $d(o_i, \mathcal{M})$ denotes the $L_2$-distance of $o_i$ to the set of vertices of the mesh $\mathcal{M}$, and where $\tau$ is a constant used to define contacts in a soft manner (\emph{i.e.} $H(o_i)=1$ when $d(o_i, \mathcal{M})=0$, and $H(o_i) \approx 0$ when $d(o_i, \mathcal{M}) >> \tau$). In our experiments, we use uniformly sampled meshes and choose $\tau=0.01m$.
Figure~\ref{fig:contact-region-heatmap} shows examples of contact heatmaps for different human grasps.

We define our object-centric contact heatmap loss as the $L_1$-distance between these generated heatmaps:
\begin{equation}
    \mathcal{L}_{C} = \sum_{ o_i \in \mathcal{M}_{object}} |H^{hand}(o_i) - H^{robot}(o_i)|
\label{eq:contact_loss}
\end{equation}

\paragraph{Hand Orientation Loss $\mathcal{L}_{O}$} %
Grasps have high similarity if the hands are oriented similarly towards the object, thus resulting in a similar free space around the object, and thus potentially to a similar affordance. Therefore, we introduce a loss to encourage the orientation of the hand and the gripper to be similar. To this end, for each human/robot hand model, we define two unit vectors which are inherent to the model, the forward vector $f$ and the normal vector $n$. 
Examples of these two vectors for different models are shown in Figure~\ref{fig:comparison}.
The normal vector $n$ is defined as the unit normal vector of the palm surface. The forward vector $f$ is defined as the unit vector that is parallel to the palm surface and pointing to the `pushing' direction. 
We define the hand orientation loss as the $L_1$-distance between these two unit vectors:
\begin{equation}
    \mathcal{L}_{O} = \vert n_{robot}  - n_{hand} \vert + \vert f_{robot} - f_{hand} \vert \> 
\label{eq:label_loss}
\end{equation}

\begin{figure*}
\centering
\includegraphics[width=0.9\linewidth]{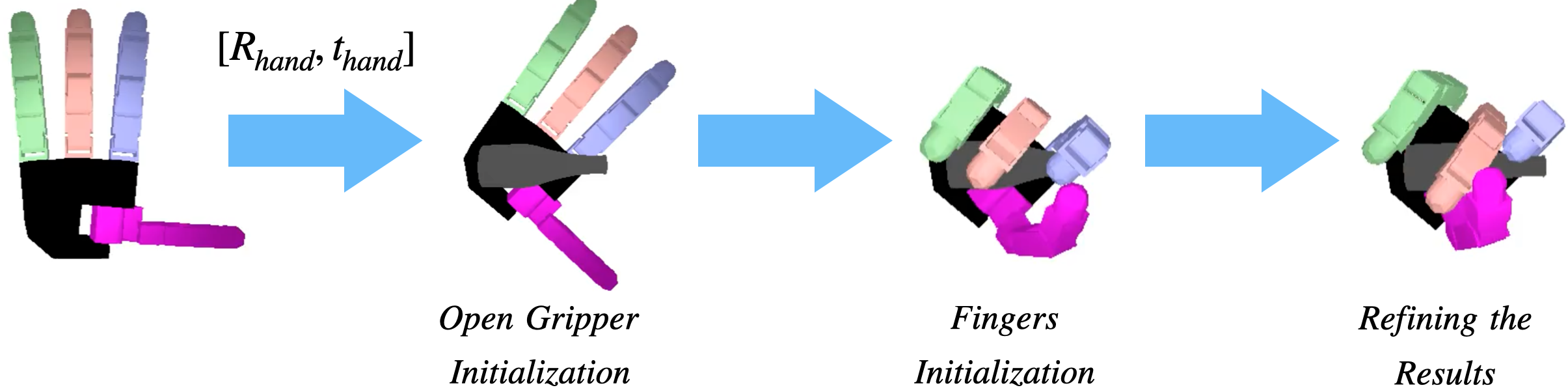}
\vspace{-0.2cm}
\caption{\label{fig:grasp-pairs-generation} \textbf{Overview of our pipeline for transferring human hand grasp to robot gripper grasp.} We first initialize the gripper with open fingers at the location of the hand. We then initialize the fingers position on the object surface by minimizing the distance between the fingertips and the contact regions of the human demonstration. At last, we refine the grasp by minimizing the overall objective function.}
\end{figure*}

\paragraph{Gripper-Object Interpenetration Loss $\mathcal{L}_{I}$}

To avoid interpenetration while ensuring realistic contacts between the robotic gripper and the object, we take inspiration from Müller~\emph{et al.}~\cite{selfcontact} and add a loss
\begin{equation}
    \mathcal{L}_{I} = \alpha_1 \mathcal{L}_{push} + \beta_1 \mathcal{L}_{pull} + \gamma_1 \mathcal{L}_{normal} \>
\label{eq:interpenetration_loss}
\end{equation}
to our objective function. 
It consists of three weighted terms.
\newline\noindent $\bullet$ The first term $\mathcal{L}_{push}$ aims at avoiding interpenetration by pushing the penetrated parts of the robotic gripper towards their nearest surface on the object mesh. To do so, we consider $\mathcal{U}_{robot} \subset \mathcal{M}_{robot}$ the set of vertices on the robotic gripper mesh that are inside the object mesh, and $\mathcal{U}_{object} \subset \mathcal{M}_{object}$ the set of vertices on the object mesh that are inside the robotic gripper mesh. In practice, we detect these two sets of vertices using the generalized winding numbers~\cite{winding_numbers}.
We define
{
\small
\begin{equation}
    \mathcal{L}_{push} = 
    \hspace{-1em}
    \sum_{o_i \in
    \mathcal{U}_{object}}
    \hspace{-1em}
    \text{tanh} \left( \frac{d(o_i, \mathcal{M}_{robot})}{\alpha_2} \right) +
    \hspace{-1em}
    \sum_{r_k \in \mathcal{U}_{robot}}
    \hspace{-1em}
    \text{tanh} \left( \frac{d(r_k, \mathcal{M}_{object})}{\alpha_2} \right)
\end{equation} 
}
to penalize interpenetration.
\newline\noindent $\bullet$
The second term $\mathcal{L}_{pull}$ encourages contacts for points of the gripper closer than a threshold $\delta$ to the object, while being constant for points farther away:
\begin{equation}
    \mathcal{L}_{pull} = \sum_{ r_k \in \mathcal{M}_{robot}} \text{tanh} \left( \frac{\min(d(r_k, \mathcal{M}_{object}), \delta)}{\beta_2} \right) %
\end{equation}
We use $\delta=2mm$ in practice.
\newline\noindent $\bullet$
To further ensure realistic contacts, a third term is added that encourages normals of both meshes to be opposite at contact locations $\mathcal{V} = \{ r_k \in \mathcal{M}_{robot} \vert d(r_k, \mathcal{M}_{object}) < \delta \}$:
\begin{equation}
    \mathcal{L}_{normal} = \sum_{r_k \in \mathcal{V}} 1 +  \langle N(r_k), N( o^k_i ) \rangle
\end{equation}
where $N(\cdot)$ denotes the unit normal vector at a given vertex, and $o^{k}_i = arg\,min_{o_i \in \mathcal{M}_{object}} d(r_k, o_i)$ denotes the closest point on the object for any vertex $r_k \in \mathcal{V}$.
Hyperparameters values are experimentally set to $\alpha_1=2.4$, $\beta_1=7$, $\gamma_1=0.001$, $\alpha_2=4cm$, $\beta_2=6cm$.

\paragraph{Gripper Self-penetration Loss $\mathcal{L}_{S}$}

$\mathcal{L}_I$ considers griper-object penetration, but some  
configurations of the gripper could also lead to self-penetration between different gripper components such as its fingers. We thus add a loss to avoid self-penetration. To this end, we use the exact same loss as $\mathcal{L}_{push}$
but apply it between the gripper mesh and itself, resulting in a loss $\mathcal{L}_{S}$.

\subsection{Optimization Pipeline}
\label{sub:pipeline}

Our objective function of Equation~\eqref{eq:objective_function} admits many local minima, and several optimization terms admit zero gradient when the gripper is far from the object. Having a good initialization is therefore important, and we thus propose a multi-step optimization pipeline whose overview is shown in Figure~\ref{fig:grasp-pairs-generation}.
It consists of 3 steps: (a) initializing the robotic gripper with open fingers around the same location as the human hand, (b) closing the fingers until contact with object, (c)  refining all degrees of freedom.

\paragraph*{(a) Open Gripper Initialization}
Because of the hand orientation loss $\mathcal{L}_O$,
the optimal global position and orientation of the gripper ($R_{robot}, t_{robot}$) is likely to be close to the global position and orientation of the human hand ($R_{hand}, t_{hand}$) in the object coordinate system. This is why we initialize the gripper position and orientation at the same position and orientation as the human hand. At this stage, we assume that the rest of the parameters, \emph{i.e.}, the angle of the finger joints correspond to a fully-open position and we thus refer to this stage as `open gripper initialization'.

\paragraph*{(b) Fingers Initialization}
To initialize the fingers and make the fingers touch the object at the right place, we first detect the contact region of the human grasp, then we minimize the distance between the fingertips and their nearest contact region using the gripper-object interpenetration loss $\mathcal{L}_{I}$ defined in Equation (\ref{eq:interpenetration_loss}) with the self-penetration loss $\mathcal{L}_{S}$.
In this way, we can put the fingers of the robot hand to their closest region of contact and at the same time avoid gripper-object interpenetration and self-penetration. 

\paragraph*{(c) Refining the Results}

We finally run the full optimization from this initialization. We use AdamW~\cite{adamw} as our optimizer, the initial learning rate is set to 0.001 for the translation $T_{robot}$ and 0.01 for rotation $R_{robot}$ and pose parameters $\theta_{robot}$. Each grasp is optimized for 100 iterations. The learning rate decreases by 10 at iteration \#50. During the optimization, we use the rotation parametrization introduced in~\cite{zhou2019continuity}.

\begin{table*}
  \centering
  \caption{\label{tab:sota} \textbf{Comparison of our approach with state-of-the-art methods.}
  $^\dagger$ indicates methods that use different hyperparameters for different grippers.
  For GraspIt!, we generated 100 grasps per human demonstration and selected the one with the lowest orientation difference and contact heatmap difference, \emph{i.e.}, the lowest $\mathcal{L}_{C}+\mathcal{L}_{O}$ loss.
  } 
  \small
   \vspace{-0.2cm}
   \begin{tabular}{ll @{\hskip 1cm} ccccc}
   \toprule
&                          & Grasp & Max Penetration & Penetration & Orientation & Contact Heatmap \\
&                          & $\epsilon$-quality $\uparrow$ & Depth ($cm$)$\downarrow$ & Volume ($cm^3$)$\downarrow$ & Difference $\downarrow$ & Difference $\downarrow$ \\
   \midrule
\multirow{4}{*}{\rotatebox[origin=c]{90}{\parbox{1cm}{\centering Allegro \\ Hand}}}
&   DexPilot$^\dagger$~\cite{dexpilot}     & {\bf 0.535} & 2.91 & 5.04 & 0.011 & 0.176 \\
&   ContactGrasp$^\dagger$~\cite{contactgrasp} & 0.460 & 3.53 & 6.94 & 1.818 & 0.195 \\
&   GraspIt! (best $\mathcal{L}_{C}+\mathcal{L}_{O}$)~\cite{graspit} & 0.345 & 2.76 & {\bf 1.27} & 0.420 & 0.254
   \\
&   \textbf{Ours}          & 0.466 & {\bf 2.57} & 4.89 & {\bf 0.001} & {\bf 0.153} \\
\midrule
\multirow{3}{*}{\rotatebox[origin=c]{90}{\parbox{0.8cm}{\centering Barrett \\ Hand}}}
&   ContactGrasp$^\dagger$~\cite{contactgrasp} & 0.523 & 4.65 & 6.28 & 2.003 & 0.225 \\
&   GraspIt! (best $\mathcal{L}_{C}+\mathcal{L}_{O}$)~\cite{graspit} & 0.354 & 4.52 & {\bf 0.88} & 0.714 & 0.258
   \\
&   \textbf{Ours}          & {\bf 0.566} & {\bf 4.09} & 2.91 & {\bf 0.001} & {\bf 0.166} \\
   \bottomrule
   \end{tabular}
\end{table*}

\section{EXPERIMENTS}
\label{sec:xp}

In this section, after presenting datasets and metrics (Section~\ref{sub:xpdata}), we provide the results of an ablation study in Section~\ref{sub:ablations} and a comparison to the state of the art in Section~\ref{sub:sota}. We then describe a user study (Section~\ref{sub:userstudy}) that validates that our approach leads to grasps more similar to the human demonstrations than the state of the art. 

\subsection{Datasets and Metrics}
\label{sub:xpdata}

To measure performance, we consider the human grasps from the YCB-Affordance dataset~\cite{ganhand}. For the 52 objects of the YCB-Objects~\cite{ycbobjects}, different types of human grasps are manually annotated and refined using GraspIt!~\cite{graspit}. This leads to a diversity in terms of grasps, including not only power grasps but also pinch grasps, \emph{etc.}, as shown by the annotations of the grasp categories provided with the dataset~\cite{ganhand} and illustrated in the top row of Figure~\ref{fig:results_examples}.

For evaluation, we measure both the grasps quality using the Grasp $\epsilon$-quality metric, which corresponds to the radius of the largest ball centered at the origin which can be enclosed by the convex hull of the wrench space~\cite{miller1999examples}, as commonly used in, \emph{e.g.}, \cite{graspit_score, multifingan, ycbobjects}. We also report the \emph{Max Penetration Depth}, \emph{i.e.}, the maximum distance between a vertex of the gripper that is inside the object and its closest vertex on the object, and the \emph{Penetration Volume}, \emph{i.e.}, the estimated volume of penetration between the two meshes.
In addition to these grasp quality metrics, we propose two metrics that are used to measure the similarity between the human hand grasp and the robot gripper grasp: the \emph{Contact Heatmap Difference} that measures contact similarity and the \emph{Orientation Difference} that measures similarity in terms of contact angle. We use the metric $\mathcal{L}_{C}$ introduced in Section~\ref{sub:loss} to evaluate numerically the \emph{Contact Heatmap Difference} and $\mathcal{L}_{O}$ for the \emph{Orientation Difference}. %

\subsection{Ablations}
\label{sub:ablations}

We first ablate our approach in Table~\ref{tab:losses} for the Allegro gripper.
To start with, we replace the second step of our optimization pipeline by another finger closing strategy inspired by~\cite{ganhand}:
starting from an \emph{open} configuration of the gripper, we discretize the gripper configuration space and pick iteratively for each finger joint the bin corresponding to the most \emph{closed} configuration that does not penetrate the object.
We observe that this significantly degrades the grasp $\epsilon$-quality, the penetration volume and the contact heatmap similarity.

We then ablate the losses of the final optimization step of our approach by removing them one by one. Removing the contact similarity loss $\mathcal{L}_{C}$ significantly degrades the contact heatmap difference from 0.153 to 0.189, while also impacting negatively the grasp $\epsilon$-quality metric and the interpenetration.
Additionally removing the loss on the angle similarity $\mathcal{L}_{O}$ leads to grasps that are even less similar and leads to higher penetration volume.
Furthermore, removing the self-penetration loss $\mathcal{L}_{I}$ degrades the grasp $\epsilon$-quality even more.
We finally evaluate the performance of our approach without the global optimization (the third step of our pipeline) in the last row, to show its importance for achieving good grasps.

\begin{table*}
\centering
\small
\caption{\label{tab:losses} \textbf{Ablation study} of our approach with the Allegro gripper. In the first row, we replace the second step of our pipeline with contact optimization for the fingers' initialization by a discrete closing strategy. In the rows below, we remove the losses one by one. The last row without any loss corresponds to the absence of Step 3 of our optimization pipeline. }
\vspace{-0.2cm}
\begin{tabular}{c @{\hskip 1em} cccc @{\hskip 1em} ccccc}
\toprule
Fingers &  \multirow{2}{*}{$\mathcal{L}_{I}$} &   \multirow{2}{*}{$\mathcal{L}_{S}$} &   \multirow{2}{*}{$\mathcal{L}_{O}$} &   \multirow{2}{*}{$\mathcal{L}_{C}$} &
                        Grasp & Max Penetration & Penetration & Orientation & Contact Heatmap \\
Init. (step 2) & & & & &
                        $\epsilon$-quality $\uparrow$ & Depth ($cm$)$\downarrow$ & Volume ($cm^3$)$\downarrow$ & Difference $\downarrow$ & Difference $\downarrow$ \\
  
\midrule
Discrete init. & \checkmark & \checkmark & \checkmark & \checkmark & 0.294 & 2.59 & 5.83 & 0.011 & 0.213 \\
\midrule
\multirow{5}{*}{Contact optim.} &
 \checkmark & \checkmark & \checkmark & \checkmark & \bf{0.466} & 2.57 & \bf{4.89} & \bf{0.001} & \bf{0.153} \\
& \checkmark & \checkmark & \checkmark &  & 0.438 & 2.70 & 6.32 & \bf{0.001} & 0.189 \\
& \checkmark & \checkmark &  &  & 0.467 & \bf{2.55} & 6.60 & 0.041 & 0.190 \\
& \checkmark &  &  &  & 0.411 & 2.69 & 6.39 & 0.031 & 0.187 \\
\cmidrule{2-10}
 &  &  &  &  & 0.387 & 2.96 & 8.52 & 0.011 & 0.170 \\
\bottomrule
\end{tabular}
\end{table*}

\subsection{Comparison with the state of the art}
\label{sub:sota}

\begin{figure*}
\centering
\small
\includegraphics[width=\linewidth]{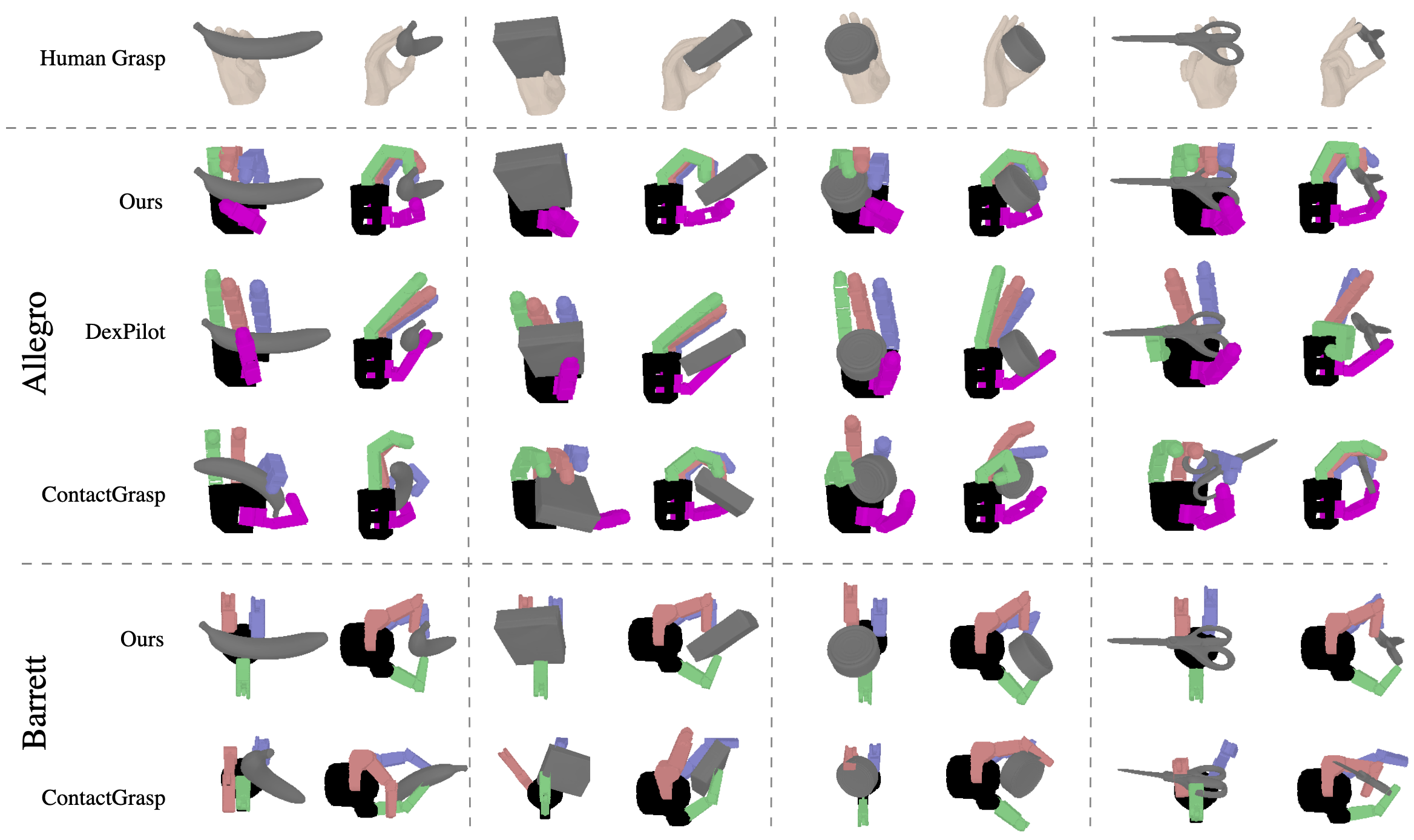}
\vspace{-0.4cm}
\caption{\label{fig:results_examples} \textbf{Example of generated grasp transfers} for our approach, DexPilot\cite{dexpilot} and ContactGrasp\cite{contactgrasp}}
\end{figure*}

We compare our approach to the state of the art and report performances for the Allegro gripper as well as for the BarrettHand gripper in Table~\ref{tab:sota}. We also show various examples in Figure~\ref{fig:results_examples}.

As a first approach, we compare to a manually-defined mapping between the human hand and the Allegro gripper
using a re-implementation of DexPilot~\cite{dexpilot}. Note that this approach would require new manual annotations for another type of robotic gripper.
As a second approach, we use ContactGrasp~\cite{contactgrasp} that proposes to refine and rerank the grasps generated by GraspIt! by exploiting contact information. For each object, GraspIt! generates grasps from different directions around the object (we consider 100 grasps in practice). The generated grasps are fed to the ContactGrasp pipeline with the contact region heatmaps generated using the code provided by the authors.
In the end, we consider the best-ranked robotic grasp for each reference human grasp.
As a third approach, we use GraspIt!~\cite{graspit} to generate 100 different grasps for an object and select the one that minimizes the sum of \emph{Orientation Difference} and \emph{Contact Heatmap Difference}.

While DexPilot obtains a higher $\epsilon$-quality metric than our approach with the Allegro gripper, the grasps are actually less realistic as there are larger interpenetrations with the object, as illustrated in the left example of Figure~\ref{fig:results_examples}. Additionally, the grasp similarity in terms of both orientation and contact heatmap is lower than our approach. Note also that DexPilot 
is not generic in that it has been handcrafted for the Allegro gripper.
We also compare our method to the best grasp from ContactGrasp~\cite{contactgrasp}, which uses different hyperparameters for the two grippers. We observe that our approach leads to higher quality grasps, with less interpenetrations, and with higher similarity with the input human grasps. This is particularly true for the orientation similarity as illustrated in the examples of Figure~\ref{fig:results_examples}: while ContactGrasp optimizes for similar contact regions, it does not enforce the gripper to approach the object from a similar direction, which can lead to grasps with significantly different properties than the human grasps in terms of free space around the object. Lastly, our method outperforms GraspIt! on every metric except for the penetration volume.

\begin{figure}
\centering
\small
\includegraphics[height=0.7in]{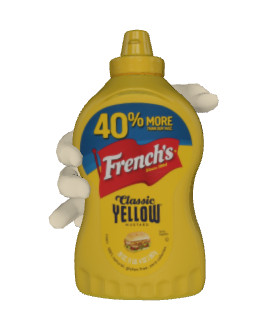}
\includegraphics[height=0.7in]{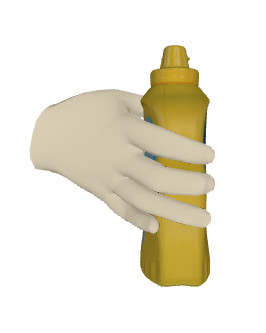}
\hfill{}
\vline{}
\hfill{}
\includegraphics[height=0.7in]{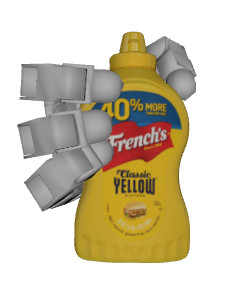}
\includegraphics[height=0.7in]{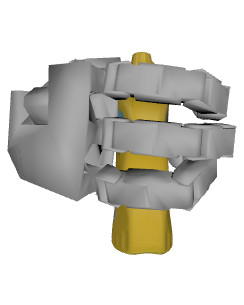}
\hfill{}
\includegraphics[height=0.74in]{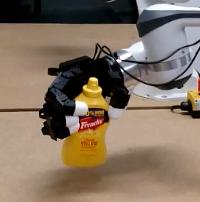}
\vspace{-0.3cm}
\caption{\label{fig:realworld} \textbf{Real-world experiments.} \textit{Left:} input human demonstration. \textit{Middle:} corresponding Allegro grasp prediction. \textit{Right:} execution.}
\end{figure}

\paragraph*{Running time.}
We evaluated all the methods on a machine with 20 Intel(R) Core(TM) i9-9900X CPUs and one NVidia GeForce RTX 2080Ti card. Our approach takes about 1 minute for a given grasp.
For comparison, our implementation of DexPilot takes about 1 second but does not consider the geometry of the object, and ContactGrasp takes on average 43 minutes for one input human grasp as it first requires to generate 100 robotic gripper grasps using GraspIt! before refining all of them.

\subsection{User Study}
\label{sub:userstudy}

We aim at enabling robots to \emph{grasp like humans}, but the metrics above do not necessarily express this notion well. We therefore conducted a user study to better evaluate the similarity between human and robotic grasps.
It is difficult for people to quantitatively evaluate this similarity, thus we resorted to a comparative evaluation.
We randomly selected 120 human grasps from the YCB-Affordance~\cite{ganhand} dataset.
For each human grasp, we generated corresponding robotic grasps using different methods and asked participants to select the one which -- in their opinion -- is the most similar to the human demonstration. 
For the Allegro gripper, we compared our method with ContactGrasp~\cite{contactgrasp} and DexPilot~\cite{dexpilot}. For BarrettHand, we compared our method with ContactGrasp~\cite{contactgrasp}. We also included an additional grasp generated randomly using GraspIt!~\cite{graspit} as baseline.

 We received in total 1,392 votes from 58 participants -- each participant sharing its preference regarding 24 human grasps. Results are summarized in Table~\ref{tab:user_study}.
Overall, the participants favored grasps produced by our method in 51\% of the cases for the Allegro gripper, and 73\% of the cases for BarrettHand. These scores are way above random chance (25\% for Allegro, 33\% for BarrettHand), and they suggest that our generated grasps are considered significantly more similar to the human demonstrations than the grasps generated using the other evaluated methods.
Further analysis of the results showed that the preference for our method could be explained in all cases by the smaller difference of global orientation between the robot and human hands when using our method.

\begin{table}
\centering
\small
\caption{\label{tab:user_study} \textbf{User-Study:}  
``Which grasp is the most similar to the human demonstration?''}
\begin{tabular}{lcc}
\toprule{}
\multirow{2}{*}{Grasp generation method} & \multicolumn{2}{c}{Number of votes (total: 1392)} \\
\cmidrule(lr){2-3}
& Allegro & BarrettHand \\
\midrule
GraspIt! (random)~\cite{graspit} & 49 & 38 \\
ContactGrasp~\cite{contactgrasp} & 117 & 101 \\
DexPilot~\cite{dexpilot} & 265 & - \\
\textbf{Ours} & \textbf{440} & \textbf{383} \\
\bottomrule{}
\end{tabular}

\vspace{-0.4cm}

\end{table}

\subsection{Real World Experiments}
\label{sub:video}

We focus in this work on predicting static grasps that describe the pose and joints configuration of a robotic gripper with respect to an object.
Grasping however is fundamentally a dynamic process, involving robot motion and contact forces.
To demonstrate the usability of our approach in real scenarios, we performed grasping experiments using an \emph{Allegro} gripper mounted on a \emph{Panda} robotic arm, from \emph{Franka Emika} (see one grasp example in Figure~\ref{fig:realworld}, and the attached video with 5 grasping examples on 5 different objects).
These experiments allowed to check that grasps produced by our method are physically feasible, while being similar to the human demonstrations.
Robot perception is out of the scope of this study, therefore we used as input some human demonstrations from the YCB-Affordance~\cite{ycbobjects} dataset, and we manually placed the objects in known poses before attempting the grasps with the robot. We did not conduct any quantitative evaluations because of this manual step.
Note that state-of-the-art methods for object~\cite{labbe2020} and hand+object pose estimation~\cite{obman,honnotate,egohands,dexycb,grab,hasson20_handobjectconsist} could be used to overcome this limitation.

\section{CONCLUSIONS}

We propose a multi-step optimization-based approach for transferring grasps from a human demonstration to a multi-fingered robotic gripper, so as to enable a robot to \emph{grasp like a human}.
The proposed approach is generic and can be applied to arbitrary multi-fingered gripper, as shown by our experimental evaluation with both Allegro and BarrettHand grippers.
Our results -- based on quantitative metrics and a qualitative user study -- suggest that it produces grasps significantly more similar to the human demonstrations than state-of-the-art methods, and we validated its applicability in the real world using an Allegro gripper mounted on a Panda arm.

\bibliographystyle{IEEEtran}
\bibliography{IEEEabrv,biblio}

\end{document}